\documentclass[10pt]{article}

\usepackage[table]{xcolor}
\usepackage{graphicx}
\usepackage{hyperref}
\usepackage{xspace}

\hypersetup{colorlinks,linkcolor={red!50!black},citecolor={blue!50!black},urlcolor={blue!80!black}}
\usepackage[toc,page]{appendix}
\usepackage[top=1.2in, bottom=1.22in, left=1.2in, right=1.2in]{geometry}

\definecolor{colorse}{RGB}{255,248,220}
\definecolor{dgreen}{RGB}{47,135,7}
\definecolor{colorex}{RGB}{255,248,220}
\definecolor{dred}{RGB}{120,6,54}

\usepackage[tikz]{bclogo}
\usetikzlibrary{shapes,calc,positioning,automata,arrows,trees}
\tikzset{
  dirtree/.style={
    grow via three points={one child at (0.8,-0.7) and two children at (0.8,-0.7) and (0.8,-1.45)}, 
    edge from parent path={($(\tikzparentnode\tikzparentanchor)+(.4cm,0cm)$) |- (\tikzchildnode\tikzchildanchor)}, growth parent anchor=west, parent anchor=south west},
}

\usepackage{listingsutf8}
\lstnewenvironment{java}{\lstset{language=java, commentstyle=\color{dgreen}\ttfamily\footnotesize, frame=single,
    framesep=5pt, basicstyle=\ttfamily\footnotesize,
    xleftmargin=5pt,xrightmargin=1pt,stringstyle=\color{dgreen}\ttfamily}}{}

\def\ace{ACE\xspace}
\def\abscon{AbsCon\xspace}
\def\x3{{\rm XCSP$^3$}\xspace}
\def\p3{{\rm PyCSP$^3$}\xspace}
\def\j3{{\rm JvCSP$^3$}\xspace}

\newcommand{\h}[1]{\texttt{#1}} 
\newcommand{\gb}[1]{{\tt #1}}

\newtheorem{remark}{Remark}

\title{\textcolor{dred}{\ace \\ A Generic Constraint Solver}} % \\ ~ \\ \includegraphics[scale=0.22]{logoAce.png} \\ ~ } 

\author{Christophe Lecoutre\\
  CRIL, University of Artois \& CNRS \\ 
  Lens, France \\ ~ ~ \\
lecoutre@cril.fr \\ ~ ~ \\ \includegraphics[scale=0.22]{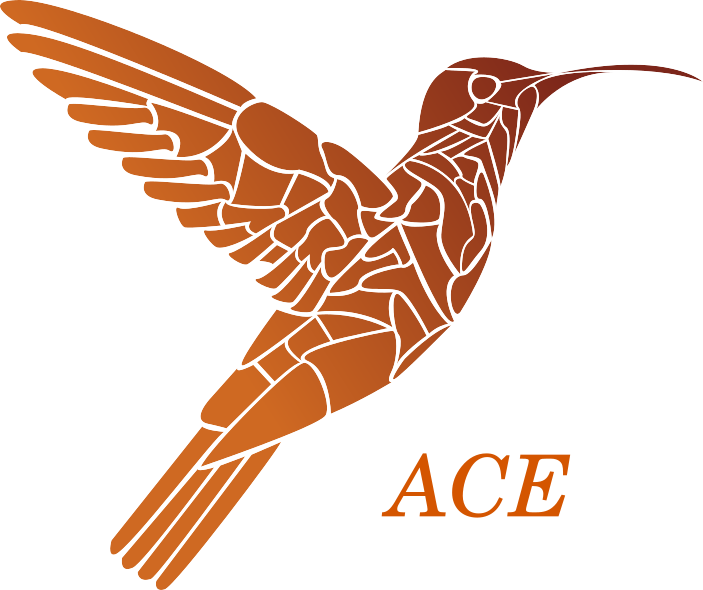}
}

\date{Version 2.4 -- August 28, 2024}

\begin{document}

\maketitle
\thispagestyle{empty}

\begin{abstract}
Constraint Programming (CP) is a useful technology for modeling and solving combinatorial constrained problems.
On the one hand, on can use a library like \p3 for easily modeling problems arising in various application fields (e.g., scheduling, planning, data-mining, cryptography, bio-informatics, organic chemistry, etc.).
Problem instances can then be directly generated from specific models and data.
On the other hand, for solving instances (notably, represented in \x3 format), one can use a constraint solver like \ace, which is presented in this paper. 
\ace is an open-source constraint solver, developed in Java, which focuses on integer variables (including 0/1-Boolean variables), state-of-the-art table constraints, popular global constraints, search heuristics
and (mono-criterion) optimization.
\end{abstract}

\section{Introduction}

Combinatorial problems are ubiquitous in the world around us.
Actually, they are found in all fields of human activity.
As illustrations, it may be a question of scheduling the operations to be carried out within an industrial process (production line of a vehicle, an airplane or a satellite), of extracting the recurring patterns in a transaction database (data mining), of organizing the roster of a service (in a hospital, university or industrial environment), of generating molecular structures with good properties (in chemistry or bioinformatics), etc.

Solving optimization problems remains a difficult task, especially when the size of the instances of the problems to be solved is large and/or when optimality is desired.
In reality, the difficulty is twofold: being able to appropriately write models for encountered problems and being able to effectively solve the different instances of these problems.
The main paradigms for optimization, namely mathematical programming, metaheuristics and Constraint Programming (CP), including the Boolean SAT framework, offer varied and interesting tools (languages, libraries, software), and are in a way, quite complementary; each paradigm having its own success stories.

Within our lab (CRIL), we have been interested for many years in Constraint Programming \cite{L_constraint}.
We have developped a Python library, \p3 \cite{pycsp3}, which allows us to write models of combinatorial constrained problems in a declarative manner.
Currently, with \p3, you can write models of constraint satisfaction and optimization problems. More specifically, you can build models for:
\begin{itemize}
\item CSP (Constraint Satisfaction Problem)
\item COP (Constraint Optimization Problem)
\end{itemize}

Once a \p3 model is written, you can compile it, while providing some data, in order to generate an \x3 instance (file), and you can solve that problem instance by means of a constraint solver which is compatible with \x3 \cite{xcsp3core,ecosystem,xcsp3}.
This is the case of solvers participating to the annual \x3 competitions (see \href{https://www.xcsp.org/competitions/}{xcsp.org/competitions} and \cite{compet22,compet23,compet24}), and of \ace in particular. 
\ace (AbsCon Essence) is derived from the constraint solver \abscon that has been used as a research platform in our team at CRIL during many years.
Many ideas and algorithms have been discarded from \abscon, so as to get a constraint solver of reasonable size and understanding.

\bigskip
Currently, we are aware of three weaknesses in \ace:
\begin{enumerate}
\item For historical reasons, the main class of \ace inherits from a class that is no longer relevant, introducing useless interfaces and code. This does not effect efficiency, but it does reduce clarity. We may decide to refactor the code in the future.  
\item \ace is bad at handling very large domains. We plan to fix this in the near future.
\item \ace lacks a good propagator for the global constraint \gb{noOverlap}. We plan to implement it (for example, following the principles of edge finding and/or not first/not last rules \cite{H_integrated}).  
\end{enumerate}

\bigskip
The paper is organized as follows.
After indicating how to run \ace, and quickly describing its main ingredients, we give some details about the 13 main packages structuring the Java code of \ace.

\section{Running \ace}

\ace is licensed under the \href{https://en.wikipedia.org/wiki/MIT_License}{MIT License}, and its code is available on \href{https://github.com/xcsp3team/ace}{GitHub}.
With the right classpath (after having cloned the code from Github), you can run the solver on any \x3 \cite{xcsp3core,ecosystem,xcsp3} instance (file) by executing:

\begin{verbatim}
  java ace <xcsp3_file> [options] 
\end{verbatim}

\noindent with:
\begin{itemize}
\item $<$xcsp3\_file$>$: an \x3 file representing a CSP or COP instance
\item $[$options$]$: possible options to be used when running the solver
\end{itemize}

\noindent Some options will be discussed later in the paper, but you can display all of them by executing:

\begin{verbatim}
  java ace
\end{verbatim}

\section{Quick Description of \ace}

\ace is an open-source constraint solver, developed in Java.
\ace focuses on:
\begin{itemize}
\item integer variables, including 0/1 (Boolean) variables
\item state-of-the-art table constraints (also called \gb{extension} constraints), including ordinary, starred, and hybrid table constraints
\item popular global constraints (\gb{allDifferent}, \gb{count}, \gb{element}, \gb{cardinality}, \gb{cumulative}, etc.)
\item search heuristics %, as e.g., wdeg \cite{BHLS_boosting,WLPT_refining}, last-conflict \cite{LSTV_reasonning}, BIVS \cite{FP_making}, solution-saving \cite{VP_ss},
\item mono-criterion optimization
\end{itemize}
It is important to note that the perimeter of \ace is basically the same as \x3-core \cite{xcsp3core}.

In this section, we first quickly describe how the central ingredients (variables, constraints and objectives) of constraint networks are coped with, in \ace, before emphasizing the importance of search, and the way it can be conducted (and controlled) in \ace.

\paragraph{Variables.}
Although many forms of variables can be represented in format \x3, as for example real, set and graph variables, \ace only deals with integer and symbolic variables.
Booolean variables are simply represented by 0/1 variables, and symbolic variables are transformed into integer variables inside the solver (just by associating arbitrary integers to symbols).
The domains of such integer variables are represented by doubly linked lists that are implemented by arrays and (partly) bit vectors, which allows us to preserve the increasing order of values.
Sparse sets, as data structure, are not used for domains, because in our experience, their drawbacks do not overcome their advantages.

Specifically, the underlying technique used behind the representation of domains corresponds to {\em dancink links}, as described in \cite{K_dancing2000,K_dancing}.
Dancing links are also used in \cite{LS_generalized}, but by the time we wrote this paper, we were not aware of the reference to the 2000 Knuth's paper.

\paragraph{Constraints.}
Generic forms of constraints, i.e., constraints represented in intension or in extension, are naturally handled in \ace.
Primitives and reification are employed for efficiently dealing with intension constraints.
Less frequent forms of intensional constraints are processed by the generic filtering scheme (G)AC3$^{rm}$ \cite{M_AC3,LH_study,LV_enforcing}.
Extension constraints, also called table constraints, are given several filtering algorithms, called propagators:  Valid-Allowed (VA) \cite{LR_fast,LS_generalized}, Simple Tabular Reduction (STR) \cite{U_partition,L_str2,LLY_str3} and Compact-Table (CT) \cite{DHLPPRS_efficiently,VLS_extending}.
Tables can be ordinary, starred, or even hybrid (smart) \cite{MDL_smart}.
All constraints handled by \ace are mainly those of \x3-core : %, and are listed in Figure \ref{fig:intCtrs}.
\begin{itemize}
\item Generic constraints: \gb{intension} and \gb{extension}
\item Language-based constraints: \gb{regular} and \gb{mdd}
\item Comparison-based constraints:  \gb{allDifferent}, \gb{allDifferentList}, \gb{allEqual}, \gb{increasing}, \gb{de\-creasing}, \gb{lexIncreasing}, \gb{lexDecreasing}, and \gb{precedence}
\item Counting constraints: \gb{sum},  \gb{count}, \gb{nValues}, and \gb{cardinality}
\item Connection constraints: \gb{maximum}, \gb{maximumArg}, \gb{minimum}, \gb{minimumArg}, \gb{element} and \gb{channel}
\item Packing and Scheduling constraints:  \gb{noOverlap},  \gb{cumulative},  \gb{binPacking} and \gb{knapsack}
\end{itemize}

\paragraph{Objectives.}
An objective is an object that must be optimized (i.e., either minimized or maximized).
This object can be a general expression (e.g., a single variable), or a value that corresponds to a computed sum, minimum, maximum or number of distinct values.
This corresponds to what is expected for mono-criterion optimization in \x3-core.
%By default, \ace uses a ramp-down strategy (that can be seen as being related to Branch\&Bound) when solving an optimization problem (i.e., a COP instance).
%Assuming a minimization problem, the principle is to add a special {\em objective constraint} $\f{obj} < \infty$ to the constraint network (although it is initially trivially satisfied), and to update the limit of this constraint whenever a new solution is found.
%It means that any time a solution $S$ is found with cost $B= \f{obj}(S)$, the objective constraint becomes $\f{obj} < B$. %, ensuring that better and better solutions are found until optimality is proved.   
%A Constraint Optimization Problem (COP) is a CSP where one objective variable has to be either minimized or maximized. 
%Thanks to a depth-first B\&B search, it is solved as a standard CSP in which a new constraint is added upon each solution to state that the next solution should be strictly better. 
%Hence, a sequence of better and better solutions is generated (satisfiability is systematically proved with respect to the current limit of the objective constraint) until no more exists (unsatisfiability is eventually proved with respect to the limit imposed by the last found solution), guaranteeing that the last found solution is optimal.

\paragraph{Search.}

\ace is a complete solver, performing a depth-first exploration of the search space, with backtracking.
At each step (node of the search tree), a decision is  taken (a variable assignment or a value refutation) and a filtering process is run (called constraint propagation).
To address the issue of heavy-tailed runtime distributions \cite{GSCK_heavy}, the search is restarted regularly, following a geometric progression (or the Luby sequence).
The order in which variables are chosen during the depth-first traversal of the search space is decided by
a variable ordering heuristic; a classical generic heuristic is \texttt{dom/wdeg} \cite{BHLS_boosting}, combined with a mechanism simulating a certain form of intelligent backjumps  \cite{LSTV_reasonning}.
The order in which values are chosen when assigning variables is decided by a value ordering heuristic; for COP instances, it is highly recommended to use first the value present in the last found solution, which is a technique known as solution(-based phase) saving \cite{VP_ss,DCS_ss}.

Backtrack search for COP relies on an optimization strategy based on decreasingly updating the maximal bound (assuming minimization) whenever a solution is found; this is a kind of ramp-down strategy (related to Branch and Bound),
whose principle is equivalent (still assuming a minimization problem) to adding a special objective constraint $\mathtt{obj} < \infty$ to the constraint
network (although it is initially trivially satisfied), and to update the limit of this
constraint whenever a new solution is found. It means that any time a solution
S is found with cost $B = \mathtt{obj}(S)$, the objective constraint becomes $\mathtt{obj} < B$.
Hence, this ramp-down strategy provides a sequence of better and better solutions until no more exist, guaranteeing that the last found solution is optimal.

At the time of writing this document, we believe that \ace is a competitive solver because of the following ingredients:
\begin{itemize}
\item restarting search frequently
\item recording nogoods from restarts \cite{LSTV_recording}
\item using constraint weighting for selecting variables \cite{BHLS_boosting,HT_conflict,WLPT_refining,HMZ_failure,ALP_guiding}
\item reasoning from last conflict(s) \cite{LSTV_reasonning}
\item using solution(-based) phase saving \cite{VP_ss,DCS_ss}
\item using BIVS \cite{FP_making} for certain problems
\end{itemize}

\begin{figure}[h]
    \centering
\scalebox{1}{\begin{tikzpicture}
\draw (0,4.5)   node{$\bullet$};
\draw (-1,3.5)   node{$\bullet$};
\draw (0,4.5) -- node [above left] {$v=a$} (-1,3.5)  ;
\draw[dashed, color=gray] (-1,3.5) -- node [above left] {$w=b$} (-2,2.5) node [below] {$\bot$} ;
\draw (-1,3.5) -- node [above right] {$w\neq b$} (0,2.5) ;
\draw (0,2.5)   node{$\bullet$};
\draw (0,2.5) -- node [ right] {$x=a$} (-1,1.5) ;
%\draw (-1,2.8) node{$\phi(P_{|x=a})$} ;
\draw (0,2.5) -- node [ right] {$x=a$} (-1,1.5) node[below] {$\bot$};

\draw (4, 3.8) rectangle node[pos=0.5] {\scriptsize Queue} (5,3.5);

\draw[->,>=latex] (-0.7,1.4) to[out=335,in=180] node[below,sloped] {$\phi(P_{|x=a})$} (3.8,3.2);
\draw[->,>=latex] (10.3,1) to[out=210,in=320] (-1,1);

\draw[fill=yellow, draw=yellow, text=black] (-1,2.1) circle (0.25) node {\bf E};
\draw[fill=orange, draw=orange, text=black] (7.3,3.9) circle (0.25) node {\bf M};
\draw[fill=red, draw=red, text=black] (11,1.6) circle (0.25) node {\bf L};

\draw (4, 3.5) rectangle  (10.7, 2.9) ;
%node[pos=.5,align=left] {$x, y, w, z, y\ldots w$}  
\draw (4.3, 3.2) node {$x$}  (5.3, 3.2) node {$y$} (6.3, 3.2) node {\color{lightgray}{$w$}} (7.3, 3.2) node {$z$} (8.3, 3.2) node  {$y$} (9.3,3.2) node{$\ldots$} (10.3, 3.2) node{{$w$}};

% contraintes x
\draw (4.3, 2.9) -- (4.3, 2.5) (4.3, 2.2) node {\color{lightgray}{$c_x^1$}} (4.3, 1.6) node {$c_x^2$};
\draw[rounded corners] (4, 2.5) rectangle (4.6,1.2);

% contraintes y
\draw (5.3, 2.9) -- (5.3, 2.5) (5.3, 2.2) node {$c_y^1$} (5.3, 1.6) node {\color{lightgray}{$c_y^2$}}   (5.3, 0.9) node {{$c_y^3$}} ;
\draw[rounded corners] (5, 2.5) rectangle (5.6,0.5);

% contraintes w
\draw (6.3, 2.9) -- (6.3, 2.5) (6.3, 2.2) node {\color{lightgray}{$c_w^1$}} (6.3, 1.6) node {\color{lightgray}{$c_w^2$}};
\draw[rounded corners] (6, 2.5) rectangle (6.6,1.2);

% contraintes z
\draw (7.3, 2.9) -- (7.3, 2.5) (7.3, 2.2) node {$c_z^1$};
\draw[rounded corners] (7, 2.5) rectangle (7.6,1.8);

% contraintes y
\draw (8.3, 2.9) -- (8.3, 2.5) (8.3, 2.2) node {\color{lightgray}{$c_y^1$}} (8.3, 1.6) node {$c_y^2$}   (8.3,  0.9) node {\color{lightgray}{$c_y^3$}} ;
\draw[rounded corners] (8, 2.5) rectangle (8.6,0.5);

% contraintes w
\draw (10.3, 2.9) -- (10.3, 2.5) (10.3, 2.2) node {\color{lightgray}{$c_w^1$}} (10.3, 1.6) node {{$c_w^2$}};
\draw[rounded corners] (10, 2.5) rectangle (10.6,1.2);

\end{tikzpicture}
}
    \caption{Illustration of pivotal moments for collecting information about conflicts: this correspond to early (E), midway (M) and late (L) processing of conflicts}
    \label{fig:weights}
\end{figure}
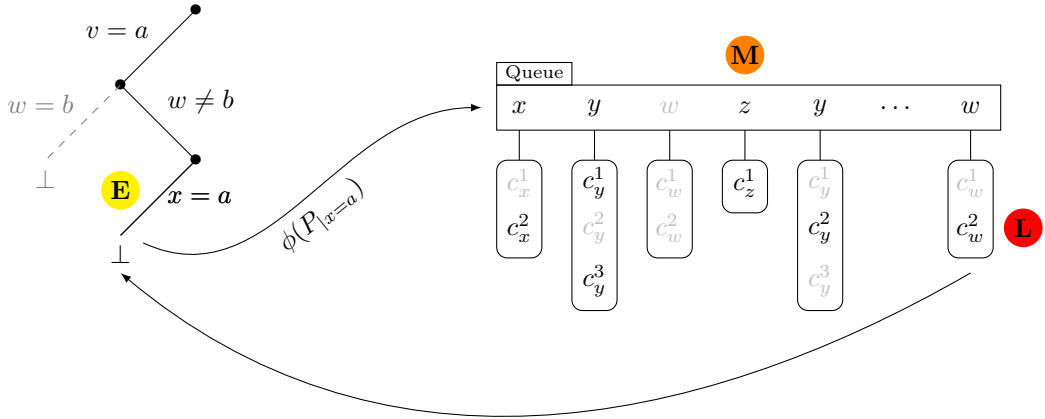

Importantly, in \cite{ALP_guiding}, we showed that three different ways of exploiting conflicts for guiding search show somewhat complementary behaviors.
This can be explained by the fact that information is extracted at different moments: at the very beginning of the process conducting to a conflict (i.e., at the time of the decision), during constraint propagation, or at the time the last propagator (filtering algorithm) is solicited. %ibiting failure).
One can then refer to such approaches as {\em early} (E), {\em midway} (M) and {\em late} (L) operational treatment of conflicts.
This is illustrated in Figure \ref{fig:weights} where a new decision $x=a$ is taken, when solving a constraint network $P$, in the continuity of two previously taken decisions $v=a$ and $w \neq b$.
In our scenario, running constraint propagation $\phi$ on (the current state of) $P$ after having assigned the value $a$ to $x$, i.e., $\phi(P|x=a)$, reveals a new conflicting (dead-end) situation (denoted by $\bot$).
The early processing of this new conflict consists in considering the variable $x$ involved in the decision as the main culprit.
This is the principle behind the heuristic \h{frba/dom} \cite{HMZ_failure}.
The midway processing of this conflict consists in considering all variables having played a role (i.e., having been picked) during propagation as having contributed to the failure.
This is the principle underlying the heuristic \h{pick/dom} \cite{ALP_guiding}.
The late processing of this conflict consists in considering the last constraint (here, $c_w^2$) provoking a domain-wipeout (i.e., removing the last value of a domain) as the object of interest.
This is the principle of constraint weighting, as in \h{wdeg/dom} \cite{BHLS_boosting,HT_conflict,WLPT_refining}.

The interest (complementary) of these three main heuristics is demonstrated at the 2024 \x3 competition \cite{compet24}, in which two versions of \ace participated:
\begin{itemize}
\item \ace, with its default behaviour (a single variable-ordering heuristic is used, \h{wdeg$^{cacd}$} \cite{WLPT_refining}),
\item \ace-mix, which corresponds to the use of the option \verb!-rr!, making use of the three main variable ordering heuristics mentionned above.
\end{itemize}

Clearly, \ace-mix benefits from the complementary of these conflict-based heuristics as this version gets better scores than \ace in the various tracks of the competition.

On the other hand, we do believe that the current trend in the CP community to widely use SAT technology (i.e., translating constraints and/or reasoning from clauses) is interesting (and has been shown to be effective in a number of situations) but somewhat detracts from the value of CP.
Regarding optimization problems, on the website showing the results of the 2024 \x3 competition (see \href{https://www.cril.univ-artois.fr/XCSP24/}{www.cril.univ-artois.fr/XCSP24}), you can see two cactus plots: 
\begin{itemize}
\item a cactus plot that gives a proof-oriented view of the results: SAT-based approaches are very efficient for proving optimality,
\item a cactus plot that gives a search-oriented view of the results (i.e., optimality proofs are discarded): CP-based approaches are very efficient for finding good quality bounds.
\end{itemize}

Actually, our position is that {\em search is reborn}, a reference to the invited talk \cite{S_dead} at CP'13 by Peter Stuckey who claimed that ``search is dead, long live proof''.
Because we face three stages when solving an optimization problem:
\begin{itemize}
\item the {\em hook phase} when a first solution must be found,
\item the {\em descent phase} in which an optimal solution must be found, along a possibly long path from the first solution,  
\item the {\em proof phase} when the proof of optimality must be performed,
\end{itemize}
it seems to us that focusing only on proof from the start is not necessarily the right approach, especially with the new advances made in search (notably, solution saving and the three complementary conflict-based heuristics).
For the descent phase, the search as performed in a classical CP solver remains extremely competitive (as shown by \ace-mix).

\section{General Packages}

The code of ACE is structured in 13 packages.

\subsection{Package \h{main}}

The package \h{main} contains two classes:
\begin{itemize}
\item \h{Head}, this is the class of the main object (head) in charge of solving a CSP or COP instance.
Such an object solicits a solver in order to solve a problem.
\item \h{HeadExtraction}, this is the class of the main object in charge of extracting an unsatisfiable core of constraints.
\end{itemize}

To run the solver on the problem instance airTraffic.xml, we can execute:
\begin{verbatim}
java main.Head airTraffic.xml
\end{verbatim}

A shortcut exists, as a class is present in the default package:
\begin{verbatim}
java ace airTraffic.xml
\end{verbatim}

To extract an unsatisfiable core of constraints from the (unsatisfiable) instance Rlfap-scen-11-f08.xml,
we execute:
\begin{verbatim}
java main.HeadExtraction Rlfap-scen-11-f08.xml
\end{verbatim}

\subsection{Package \h{dashboard}}
  
In the package \h{dashboard}, we find the following three classes:
\begin{itemize}
\item \h{Control}, which allows us to manage all options concerning the problem (typically, the way to
represent it) and the solver (typically, the way to conduct search).
\item \h{Input}, which allows us to handle arguments given by the user on the command line. These
arguments may concern the problem to solve or more generally the solving process (i.e., options
to choose like for example which search heuristics to use).
\item \h{output}, whose role is to output some data/information concerning the solving process of problem
instances. These data are collected by means of an XML document that may be saved. A part
of the data are also directly displayed on the standard output.
\end{itemize}

The class \h{Contro}l is rather central, as it contains many important options. The user can specify
the value of these options on the command line. As an example, in an intern class (representing an
option group) of the class Control, we find:

\begin{verbatim}
long timeout = addL("timeout", "t", PLUS_INFINITY, "The limit in milliseconds
                        before stopping; seconds can be indicated as in -t=10s");
\end{verbatim}

The second parameter of the method \h{addL} is the shortcut to employ on the command line. So,
to run the solver on the problem instance airTraffic.xml during at most 10 seconds, we need to
execute:
\begin{verbatim}
java ace airTraffic.xml -t=10s
\end{verbatim}

All options are briefly introduced, when executing: \verb!java ace!

\subsection{Package \h{interfaces}}

It is important to understand the role of the Java interfaces in the code of ACE.

\subsubsection{Tags}
Some interfaces are used as tags (i.e., they do not contain any piece of code):
\begin{itemize}
\item \h{TagSymmetric}, tag for indicating that a constraint is completely symmetric
\item \h{TagNotSymmetric}, tag for indicating that a constraint is not symmetric at all
\item \h{TagCallCompleteFiltering}, tag for indicating that a constraint can produce full filtering at each call (not only around the last touched variable)
\item \h{TagAC}, tag for indicating that a constraint guarantees enforcing (G)AC
\item \h{TagNotAC}, tag for indicating that a constraint does not guarantee enforcing (G)AC
\item \h{TagNegative}, tag for indicating that a table constraint is negative (i.e., contains conflicts)
\item \h{TagPositive}, tag for indicating that a table constraint is positive (i.e., contains supports)
\item \h{TagStarredCompatible}, tag for indicating that a constraint may contain starred elements (i.e., may contain *)
\item \h{TagMaximize}, tag for indicating that an object (e.g., an heuristic) aims at maximizing an expression (variable, sum, maximum, etc.)
\end{itemize}

\subsubsection{Observers}

Some interfaces are used for observing.
\begin{itemize}
\item \h{ObserverOnConstruction}, for observing the construction of the main objects (problem and solver). Methods are:
  \begin{itemize}
  \item \h{beforeAnyConstruction}, called before the main objects (problem and solver) are started to be built
  \item \h{afterProblemConstruction}, called when the construction of the problem is finished
  \item \h{afterSolverConstruction}, called when the construction of the solver is finished
  \end{itemize}
\item \h{ObserverOnSolving}, for observing the main steps of solving (preprocessing and search). Methods are:
  \begin{itemize}
  \item \h{beforeSolving}, called before solving (preprocessing followed by search) is started
  \item \h{beforePreprocessing}, called before preprocessing is started
  \item \h{afterPreprocessing}, called after preprocessing is finished
  \item \h{beforeSearch}, called before (backtrack) search is started
  \item \h{afterSearch}, called after (backtrack) search is finished
  \item \h{afterSolving}, called after solving (preprocessing followed by search) is finished
  \end{itemize}
\item \h{ObserverOnRuns}, for observing successive runs (possibly, only one run if restarting is deactivated) performed by the solver. Methods are:
  \begin{itemize}
  \item \h{beforeRun}, called before the next run is started
  \item \h{afterRun}, called after the current run is finished
  \end{itemize}
\item \h{ObserverOnBacktracks}, for observing backtracks performed by the solver. Methods and sub-interfaces are:
  \begin{itemize}
  \item \h{restoreBefore(int depthBeforeBacktrack)}, called when a restoration is required due to a backtrack coming from the specified depth
  \item interface \h{ObserverOnBacktracksSystematic}, for observing backtracks performed by the solver. Used for observers that systematically require restoration
  \item interface \h{ObserverOnBacktracksUnsystematic}, for observing backtracks performed by the solver. Used for observers that does not systematically require restoration.
  \end{itemize}
\item \h{ObserverOnDecisions}, for observing decisions taken by the solver. Methods are:
  \begin{itemize}
  \item \h{beforePositiveDecision(Variable x, int a)}, called when a positive decision (variable assignment $x=a$) is going to be taken
  \item \h{beforeNegativeDecision(Variable x, int a)}, called when a negative decision (value refutation $x \neq $) is going to be taken
  \end{itemize}
\item \h{ObserverOnAssignments}, for observing assignments taken by the solver. Methods are:
  \begin{itemize}
  \item  \h{afterAssignment(Variable x, int a)}, called after the variable assignment $x=a$ has been taken
  \item \h{afterUnassignment(Variable x)}, called after the variable $x$ has been unassigned (due to backtrack)
  \end{itemize}
\item \h{ObserverOnRemovals}, for observing reductions of domains (typically, when filtering) by the solver. Methods are:
  \begin{itemize}
  \item \h{afterRemoval(Variable x, int a)}, called when the index $a$ (of a value) for the domain of $x$ has been removed
  \item \h{afterRemovals(Variable x, int nRemovals)}, called when the domain of the variable $x$ has been reduced; the number of deleted values is specified
  \end{itemize}
\item \h{ObserverOnConflicts}, for observing conflicts encountered during search by the solver. Methods are:
  \begin{itemize}
  \item \h{whenWipeout(Constraint c, Variable x)}, Called when the domain of the specified variable has become empty (so-called domain wipeout) when filtering with the specified constraint
  \item \h{whenBacktrack}, called when the solver is about to backtrack
  \end{itemize}
\end{itemize}

\subsubsection{Specific Filtering}
There is an important interface that is used for specifying that a constraint has its own filtering algorithm (propagator).

\begin{java}
public interface SpecificPropagator {

  /**
  * Runs the propagator (specific filtering algorithm) attached
  * to the constraint implementing this interface, and returns
  * false if an inconsistency is detected.
  * We know that the specified variable has been picked from the
  * propagation queue, and has been subject to a recent reduction
  * of its domain.
  *
  * @param evt  a variable whose domain has been reduced
  * @return false if an inconsistency is detected
  */
  boolean runPropagator(Variable evt);
}
\end{java}

\subsection{Packages \h{sets} and \h{utility}}

There are three main implementations of sets (which are important structures) in ACE, which can be
found in the package sets:
\begin{itemize}
\item \h{SetDense}, a dense set is basically composed of an array (of integers) and a limit: at any time, it contains the elements (typically, indexes of values) in the array at indexes ranging from 0 to the limit (included).
\item \h{SetSparse}, a sparse set \cite{BT_efficient} is basically composed of two arrays (of integers) and a limit: at any time, it contains the elements (typically, indexes of values) in the array ‘dense’ at indexes ranging from 0 to the limit (included). The presence of elements can be checked with the array ‘sparse’.
Sparse sets are always simple, meaning that values in ‘dense’ are necessarily indexes 0, 1, 2, ...
Besides, we have \verb!dense[sparse[value]] = value!.
\item \h{SetLinked}, a linked set is an interface that allows us to represent a list of elements perceived as indexes, i.e., elements whose values range from 0 to a specified capacity -1.
For instance, if the initial size (capacity) of the object is 10, then the list of indexes/elements is 0, 1, 2..., 9.
One can remove indexes of the list, one can iterate, in a forward or backward way, the currently present indexes, and one can iterate over deleted indexes. Each deleted index has an associated level.
This kind of interface is notably used for managing the indexes of values of variable domains.
Roughly speaking, \h{SetLinked} can be seen as an implementation of {\em dancink links}, as described in \cite{K_dancing2000,K_dancing}.
\end{itemize}

There are several subclasses:
\begin{itemize}
\item \h{SetDenseReversible}, a dense set that can be handled at different levels (of search)
\item \h{SetSparseReversible}, a sparse set that can be handled at different levels (of search)
\item \h{SetLinkedBinary} and \h{SetLinkedFinite}, for sets containing only two elements (indexes 0 and 1), and ordered sets containing a finite number of elements (indexes), respectively.
\end{itemize}

In the package \h{utility}, there are several classes with utility methods:
\begin{itemize}
\item \h{Bit}, containing methods on bit vectors
\item \h{Kit}, containing various useful methods
\item \h{Reflector}, containing methods for performing operations based on reflection
\item \h{Stopwatch}, containing methods for measuring the time taken by some operations, when solving a problem
\end{itemize}

\section{Packages concerning the Problem Representation}

\subsection{Package \h{problem}}

The package \h{problem} contains the main class \h{Problem}, as well as some auxiliary classes. The main classes of the package are:
\begin{itemize}
\item \h{Problem}, the class for representing the problem instance (constraint network)
\item \h{Features}, the class that is useful for storing various information (features such as sizes of
domains, types of constraints, etc.) about the problem instance, and ways of displaying it
\item \h{Reinforcer}, the class containing inner classes that are useful to reinforce a constraint network
by adding either redundant constraints or symmetry-breaking constraints
\item \h{XCSP3}, the class that allows us to load instances in \x3 format. This class is the interface
part while the class Problem is the implementation part. This separation is due to historical
reasons (from the API \j3), but could be removed in the future so as to simplify code.
\end{itemize}

\subsection{Package \h{variables}}

The package \h{variables} contains classes for defining both variables and domains. It also contains a
class that is useful to iterate over the tuples of the Cartesian product of specified domains. The main
classes of the package are:
\begin{itemize}
\item \h{Variable}, this is the abstract root class for any variable involved in a problem, with currently
two inner subclasses: \h{VariableInteger} and \h{VariableSymbolic}. A domain is attached to a
variable and corresponds to the (finite) set of values which can be assigned to it. When a value
is assigned to a variable, the domain of this variable is reduced to this value. When a solver tries
to assign a value to a variable, it uses a value ordering heuristic in order to determine which
value must be tried first. A variable can of course occur in different constraints of the problem
to which it is attached.
\item \h{Domain}, this is the abstract root class for the domain of any variable. A domain is composed of a
set of integer values. The domain is initially full, but typically reduced when logically reasoning
(with constraints). When handling a domain, to simplify programming, one usually iterates
over the indexes of the values; if the domains contains d values, the indexes then range from 0
to d-1. For instance, if the domain is the set of values $\{1,4,5\}$, their indexes are respectively
$\{0,1,2\}$. The correspondence between indexes of values and values is given by the methods
\h{toIdx} and \h{toVal}. Subclasses of \h{Domain} are:
\begin{itemize}
\item \h{DomainBinary}, the class for binary domains typically containing 0 and 1 only.
\item \h{DomainRange}, the class for domains composed of a list of integers included between two (integer) bounds.
\item \h{DomainValues}, the class for domains composed of a list of integers that are not necessarily contiguous. Note that the values are sorted.
\end{itemize}
Note that \h{DomainRange} and \h{DomainValues} are two inner subclasses of \h{DomainFinite}.
\item \h{TupleIterator}, the class that allows us to iterate over the Cartesian product of the domains of
a sequence of variables. For example, it allows us to find the first valid tuple, or the next tuple
that follows a specified (or last recorded) one. Each constraint is equipped with such an object.
It is important to note that iterations are performed with indexes of values, and not directly
values.
\end{itemize}

\subsection{Package \h{constraints}}

The package \h{constraints} contains many classes representing various types of constraints. Constraints can be defined:
\begin{itemize}
\item extensionally, by means of tables or diagrams
\item intensionally, by means of general Boolean tree expressions, which often correspond to classical forms called primitives
\item from a well-known template (so-called global constraints).
\end{itemize}
Consequently, the more abstract classes from the package are:
\begin{itemize}
\item \h{Constraint}, the root class of any constraint. Let us recall that a constraint is attached to a
problem, involves a subset of variables of the problem, and allows us to reason so as to filter the
search space (i.e., the domains of the variables).
\item \h{ConstraintExtension}, the root class for representing extension constraints, also called table
constraints. Two direct subclasses are ExtensionGeneric (for implementing AC filtering à
la AC3$^{rm}$) and ExtensionSpecific (for implementing specific propagators). Algorithms for
filtering extension constraints can be found in the subpackage \h{extension}:
\begin{itemize}
\item STR1, STR2, and STR3
\item CT
\item CMDD
\end{itemize}

\item \h{ConstraintIntension}, the root class for representing intension constraints, which are constraints whose semantics is given by a Boolean expression tree involving variables. Most of the
times, primitives are used instead of this general form. Algorithms for filtering primitive and
reified constraints can be found in the subpackage \h{intension}.
\item \h{ConstraintGlobal}, the root class for representing global constraints, which are essentially constraints with a specific form of filtering (propagator). Algorithms for filtering global constraints
can be found in the subpackage  \h{global}.
\end{itemize}

\section{Packages concerning the Solving Process}

\subsection{Package \h{solver}}

The package \h{solver} contains the main class \h {Solver}, as well as some auxiliary classes that are useful
for managing future variables (i.e., unassigned variables), decisions, solutions, restarts, last conflicts
and statistics. The main classes of the package are:
\begin{itemize}
\item \h{Solver}, the class of the main object used to solve a problem.
\item \h{FutureVariables}, the class of the object that allows us to manage past and future variables.
\item \h{Decisions}, the class of the object that allows us to store the set of decisions taken by the solver
during search. Each decision is currently encoded under the form of an int. However, it is
planned in the future to represent each decision with a long instead of an int.
\item \h{Solutions}, the class of the object used to record and display solutions (and bounds, etc.).
\item \h{Restarter}, the root class of the object used for manage restarts (successive runs restarting from
the root node).
\item \h{LastConflicts}, the class of the object implementing last-conflict reasoning (lc); see \cite{LSTV_reasonning}.
\item \h{Statistics}, the class of the object that allows us to gather all statistics (as e.g., the number of backtracks) of the solver.
\end{itemize}

\subsection{Package \h{propagation}}

In the package \h{propagation}, we can find classes that implement different ways of conducting the
filtering process (or constraint propagation), all inheriting from the root class Propagation. For
example, among them, we find those corresponding to the consistencies AC (Generalized Arc Con-
sistency), SAC (Singleton Arc Consistency) and GIC (Global Inverse Consistency). For simplicity,
propagation and consistency concepts are not distinguished. This is why some subclasses are given
the name of consistencies.

Among subclasses, we find:
\begin{itemize}
\item  \h{Backward}, which is the root class for backward propagation. Such form of propagation corre-
sponds to a retrospective approach that deals with assigned variables, meaning that the domains
of the unassigned variables are never modified. Subclasses are BT, the basic backtracking algo-
rithm, and GT, the “generate and test” algorithm.

\item \h{Forward}, which is the root class for forward propagation. Such form of propagation corresponds
to a prospective approach that deals with unassigned variables, meaning that the domains of
the unassigned variables can be filtered. Subclasses are:
\begin{itemize}
\item FC, the class for Forward Checking (FC)
\item AC, the class for (Generalized) Arc Consistency (AC). Such a propagation object solicits
every constraint propagator (i.e., filtering algorithm attached to a constraint) until a fixed
point is reached (contrary to FC). Note that it is only when every propagator ensures AC
that AC is really totally enforced on the full constraint network. Let us recall that AC is
the maximal level of possible filtering when constraints are treated independently. Note
that, for simplicity, we use the acronym AC (and not GAC) whatever is the arity of the
constraints.
\item GIC, the class for Global Inverse Consistency (GIC). This kind of consistency is very strong
as once enforced, it guarantees that each literal (x,a) belongs to at least one solution. It can
only be executed on some special problems. For example, see \cite{BFL_global}.
\item SAC, the class for Singleton Arc Consistency (AC). Some information about SAC and al-
  gorithms enforcing it can be found for example in \cite{BCDL_efficient}.
\end{itemize}
\end{itemize}

\subsection{Package \h{learning}}

In the package \h{learning}, we can find classes related to constraint learning. Some of them are about
nogoods, and some are about so-called Inconsistent Partial States (IPSs).

\begin{remark}
Currently, for historic and maintenance reasons, IPS learning is deactivated. We plan, in the near future, to reactivate it.
\end{remark}

Entities that can be learned during search are from classes:
\begin{itemize}
\item \h{Nogood}; strictly speaking, an object of this class is used as if it was a nogood constraint, i.e.,
a disjunction of negative decisions that must be enforced (to be true). However, nogoods are
handled apart during constraint propagation.
\item \h{Ips}; an IPS is an Inconsistent Partial State. It is composed of membership decisions and is
equivalent to generalized nogoods.
To identify and reason with such entities, we have the classes:
\item \h{NogoodReasoner}, which allows us to record and reason with nogoods.
\item \h{IPsReasoner}, which allows us to record and reason with inconsistent partial states (IPSs). It is
possible to reason in term of equivalence or dominance.
Currently, learning is limited to nogood recording from restarts.
\end{itemize}

\subsection{Package \h{heuristics}}

\subsubsection{Variable Ordering Heuristics}

Variable ordering heuristics can be found in the package heuristics, implemented by Java classes
inheriting from \h{HeuristicVariables}. Let us suppose that we want to implement the classical dynamic
variable ordering heuristic dom that selects the variable with the smallest domain size. We have to
write a class inheriting from HeuristicVariablesDynamic while specifying a body for the method
\h{scoreOf}, as follows:

\begin{java}
public class Dom extends HeuristicVariablesDynamic {
  public Dom(Solver solver, boolean anti) {
    super (solver, anti);
  }
  
  @Override
  public double scoreOf (Variable x) {
    return x.dom.size();
  }
}
\end{java}

In case the highest score must be selected (and not the smallest one, as for dom), the class must
be tagged as follows: \verb!implements TagMaximize!.
Once a new heuristic is available, we can select it by using the option -varh. For example, to run
the solver on the problem instance airTraffic.xml, while using dom as variable ordering heuristic,
we just have to execute:
\begin{verbatim}
java ace airTraffic.xml -varh=Dom
\end{verbatim}

If for some reasons, we want to use the anti-heuristic, we execute:
\begin{verbatim}
  java ace airTraffic.xml -varh=Dom -anti_varh
\end{verbatim}

Among implemented heuristics, we find:
\begin{itemize}
\item Rand
\item Dom
\item DDegOnDom
\item Wdeg
\item WdegOnDom
\item FrbaOnDom
\item PickOnDom
\item Activity and Impact, with basic implementations
\end{itemize}

For the constraint-weighting heuristics Wdeg and WdegOnDom, four variants exist:
\begin{itemize}
\item VAR, a basic variant
\item UNIT, classical weighting, as described in \cite{BHLS_boosting} 
\item CACD, refined weighting, as described in \cite{WLPT_refining} 
\item CHS, as described in \cite{HT_conflict}
\end{itemize}

By default, the variant CACD of Wdeg is selected. For specifying a variant, the option -wt must be
used. For example, to run the solver on airTraffic.xml, while using the variant CHS of WdegOnDom,
we need to execute:
\begin{verbatim}
java ace airTraffic.xml -varh=WdegOnDom -wt=chs
\end{verbatim}

To use the search strategy of \ace-mix at the 2024 \x3 competition, add the option \verb!-rr!, as in:
\begin{verbatim}
java ace airTraffic.xml -rr
\end{verbatim}

\subsubsection{Value Ordering Heuristics}

Value ordering heuristics can be found in the package heuristics, implemented by Java classes
inheriting from HeuristicValues. Let us suppose that we want to implement the value order-
ing heuristic rand that randomly selects a value in the domain of the selected variable. We have
to write a class inheriting from HeuristicValuesDirect while specifying a body for the method
\h{computeBestValueIndex}, as follows:

\begin{java}
  public class Rand extends HeuristicValuesDirect {
    public Rand (Variable x, boolean dummy) {
      super (x, dummy);
    }
    
    @Override
    public int computeBestValueIndex() {
      return x.dom.any();
    }
  }
\end{java}

Once a new heuristic is available, we can select it by using the option -valh. For example, to run
the solver on the problem instance airTraffic.xml, while using rand as value ordering heuristic, we
just have to execute:
\begin{verbatim}
java ace airTraffic.xml -valh=Rand
\end{verbatim}

Among implemented heuristics, we find:
\begin{itemize}
\item Rand
\item First and Last
\item Robin and RunRobin
\item Bivs, as defined in \cite{FP_making}
\end{itemize}

\subsection{Package \h{optimization}}

The package \h{optimization} contains classes that are useful for managing optimization. The main
classes of the package are:
\begin{itemize}
\item \h{Optimizer}, the abstract root class of the object used as a pilot for dealing with (mono-objective)
optimization. Subclasses define various strategies to conduct search toward optimality:
\begin{itemize}
\item \h{OptimizerDecreasing}, an optimization strategy based on decreasingly updating the max-
imal bound (assuming minimization) whenever a solution is found; this is related to branch
and bound (and related to ramp-down strategy)
\item \h{OptimizerIncreasing}, an optimization strategy based on increasingly updating the min-
imal bound (assuming minimization); this is sometimes called iterative optimization (or
ramp-up strategy)
\item \h{OptimizerDichotomic}, an optimization strategy based on a dichotomic reduction of the
bounding interval
\end{itemize}
\item \h{Optimizable}, the interface that any constraint must implement so as to represent an objective
\item \h{ObjectiveVariable}, the abstract root class of the object to be used when the objective of the
problem instance (constraint network) is given by a variable whose value must be minimized or
maximized. Inner subclasses are \h{ObjVarLE} and \h{ObjVarGE}.
\end{itemize}

\section*{Some Useful links}

\begin{itemize}
\item \p3 website: \href{https://www.pycsp.org}{pycsp.org}
\item \ace Github: \href{https://github.com/xcsp3team/ace}{github.com/xcsp3team/ace}
\item \x3 website: \href{https://www.xcsp.org}{xcsp.org}
\end{itemize}

\section*{Acknowledgements}
This work has been supported by the project CPER CornelIA from the ``Hauts-de-France'' Region, as well as the National Research Agency under France 2030, MAIA Project ANR-22-EXES-0009.

%\bibliographystyle{plain} %alpha}
%\bibliography{globalBiblio}

\end{document}